# Modeling and Simulation of a Multi-Robot System Architecture

Ahmed R. Sadik
*Honda Research Institute Europe*
Offenbach am Main, Germany
ahmed.sadik@honda-ri.de

Christian Goerick
*Honda Research Insitute Europe*
Offenbach am Main, Germany
christian.goerick@honda-ri.de

Manuel Muehlig
*Honda Research Insitute Europe*
Offenbach am Main, Germany
manuel.muehlig@honda-ri.de

*Abstract*—A Multi-Robot System (MRS) is the infrastructure of an intelligent cyber-physical system, where the robots understand the need of the human, and hence cooperate together to fulfill this need. Modeling an MRS is a crucial aspect of designing the proper system architecture, because this model can be used to simulate and measure the performance of the proposed architecture. However, an MRS solution architecture modeling is a very difficult problem, as it contains many dependent behaviors that dynamically change due to the current status of the overall system. In this paper, we introduce a general purpose MRS case study, where the humans initiate requests that are achieved by the available robots. These requests require different plans that use the current capabilities of the available robots. After proposing an architecture that defines the solution components, three steps are followed. First is modeling these components via Business Process Model and Notation (BPMN) language. BPMN provides a graphical notation to precisely represent the behaviors of every component, which is an essential need to model the solution. Second is to simulate these components behaviors and interaction in form of software agents. Java Agent DEvelopment (JADE) middleware has been used to develop and simulate the proposed model. JADE is based on a reactive agent approach, therefore it can dynamically represent the interaction among the solution components. Finally is to analyze the performance of the solution by defining a number of quantitative measurements, which can be obtained while simulating the system model in JADE middleware, therefore the solution can be analyzed and compared to another architecture.

## I. Introduction

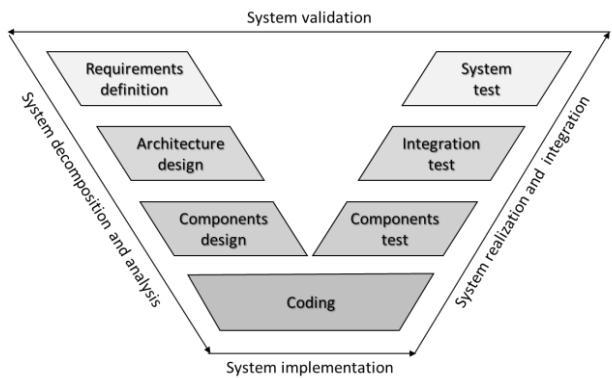

Fig. 1 simplified version of the V-model – adapted from [3]

A Multi-Robot System (MRS) is an intelligent solution for complex problems that need more than a single robot with all the required capabilities to solve this problem [1]. The MRS has gained a great focus in research since the nineties. Now, the MRS is mature enough to be one of the core technologies of Industry 4.0, and cyber-physical system. Many applications of MRS can be seen in unmanned aerial vehicles, multi-robot surveillance, search and rescue missions, services robots in smart homes and warehouses [2]. The advantages of an MRS is the time and effort efficiency, reliability and robustness. However, the problem with the MRS is the complexity of implementation. Therefore, modeling and simulation of an MRS is an important necessity to avoid this complexity.

An MRS is a typical example of a complex system, which fall under the umbrella of the system engineering field [3]. The International Council On System Engineering (INCOSE) uses the V-Model to define the necessary procedure to develop a complex system. Fig. 1 is a simplified version of the V-Model that fits the MRS development. The left side of the model focuses on the system decomposition and analysis, where the MRS architecture is built based on the system requirements. Afterwards, the system is divided into components that are modeled in details to be coded and integrated. Ultimately, during the system validation, the system is evaluated over the different design levels. Following the V-model roadmap, this article proposes an approach to model, simulate, and evaluate the performance of the MRS architecture.

As the MRS has many different applications, section II of the paper will describe a specific case study that can be used to define the system requirements. Section III introduces the fundamentals that are required to model, simulate, and evaluate the performance of the case study. Modeling the case study is explained in details in section IV, while its simulation is shown in section V. Thus, the system performance can be analyzed in section VI. Finally, section VII summarizes and discusses the work and introduces the future research.

## II. Problem and Case Study Description

An MRS is a practical application of an Information System (IS). An IS architecture is the practice of designing an IS, by defining the set of components that are required to build the system, the components structure, behaviors, and the relations and interaction among these components [4]. An IS architecture is an abstract overall depiction of the system that reflects the system functions, and constrains [5]. Therefore, it provides a graspable mean to reason and enhance the system properties and requirements, and blueprints that distingue the system concept from its implementation technology [6]. An IS performance is an important feature to evaluate the system architecture. However, there is no clear definition for quantitative performance criteria that can be measured while the MRS is running, and hence can be used to analysis the overall system behavior. Therefore, the problem of this article is to provide an approach to model, simulate and evaluate the performance of an MRS architecture,

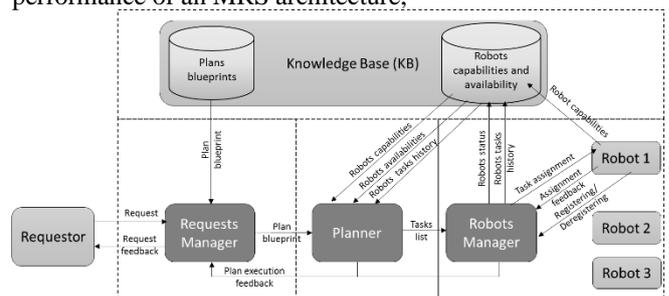

Fig. 2. case study component diagram

As an MRS involves a variety of other research topics such as planning and scheduling [7], knowledge representation [8], and reasoning [9]. The proposed MRS architecture uses the most simplified method to model these aspects, without influencing the main system functions and processes. The case study of this research is shown in Fig. 2. The proposed architecture is built upon the component-based architectural pattern, which is a common pattern for a multi-tier complex system [10]. The main components define this architecture are the requests manager, the planner, and the robots manager. The requests manager receives a specific request (Rq) from a requestor, that could be a human or a software application. The requests manager checks if the Knowledge Base (KB) contains a plan blueprint (Pb) that can fulfill the current request. A plan blueprint is a sequence of tasks (T), i.e., $Pb_i = \{T_1,...,T_n\}$, where n is the number of tasks in this blueprint and can be different from one blueprint to another. A task is a function of the capabilities (C) of the robot (R), i.e., $T_i = f(C_x, C_y,…)$, where every robot has a different set of capabilities. If the requests manager found an equivalent plan blueprint to a specific request, it forwards it to the planner. The planner checks which robots are available, and then checks if the available robots have the capabilities to achieve the tasks in the plan blueprint. If more than one robot have the capabilities to achieve a specific task, the planner compares the number of tasks that have been done by these robots in the past. Upon this comparison, the planner decides which robot will be assigned for this task. If the planner matches all the tasks of the plan blueprint with the available robots, it sends a verified plan (P) to the robots manager. The robots manager assigns the tasks of the verified plan to the robots and get the task execution feedback.

In the previous scenario, three types of variations are considered during the system running time. First is the change in the plans blueprint, by adding a new plan blueprint, deleting an existing blueprint, or modifying an existing plan blueprint. The second source of variation is the number of the available robots. This case study considered a maximum number of three robots that can exist at the time. The robots need to register to or deregister from the MRS via the robots manager. The final source of variation is the capabilities of the robots. In case of updating or modifying the robot capabilities, the robot must deregister from the MRS. Then when it registers itself again, it updates its capabilities in the KB.

### III. SOLUTION FUNDAMENTALS

#### A. Business Process Model and Notation

Business Process Model and Notation (BPMN) is an Architecture Description Language (ADL) that extends the Unified Modeling Language (UML). UML overcomes the informality of box-and-line languages [10], and the lack of mature tools of the other formal ADLs [12]. UML introduces a standard set of diagrams to visualize the different aspects and behaviors of the system. One of the most important UML diagrams is the activity diagram, which constructs a process model via standard visualization elements [13].

TABLE I. BPMN ELEMENTS THAT ARE USED IN MODELING THE CASE STUDY [15]

| Name | Type | Notations | Description |
|---|---|---|---|
| Activity (a routine or a task that happens during the process) | standard activity | | an activity that performs a general task or routine |
| | send message activity | | an activity that performance a task and then send a message to start another activity in another process |
| | receive message activity | | an activity that starts due to receiving a massage from another activity |
| Event (a happening that triggers an activity or results due to an activity) | standard start | | an event that starts an activity at the very beginning of the process |
| | standard intermediate | | an event that starts an activity during the process |
| | standard end | | an event that ends an activity at the end of the process |
| | catch message start | | a received message event that starts an activity at the very beginning of the process |
| | catch message intermediate | | a received message event that starts an activity during the process |
| | throw message intermediate | | a sent message event that happens due to an activity during the process |
| | throw message end | | a sent message event that happens due to an activity at the end of the process |
| | catch signal start | | a signal event that starts an activity at the very beginning of the process |
| | catch signal intermediate | | a signal event that starts an activity during the process |
| | throw signal end | | a signal event that happens due to an activity at the end of the process |
| | start timer | | a timer event that starts at the very beginning of the process |
| | Intermediate timer | | a timer event that starts during the process |
| Flow (the flow order of the activities) | sequential | | a sequential execution path of flow of activities |
| | default | | a default path of flow in case all the other paths are false |
| Gateway (a logic gate that controls the flow of the activities) | exclusive OR merge | | when any of the ingoing paths is true, it sets the outgoing path to be true |
| | exclusive OR split | | when the ingoing path is true, it sets only one of the outgoing path to be true |
| | inclusive OR merge | | when more than one active paths are true, it sets the outgoing path to be true |
| | inclusive OR split | | when the ingoing path is true, it sets one or more of the outgoing paths to be true |
| | parallel AND merge | | when all the ingoing paths are true, it set the outgoing path to be true |
| | parallel AND split | | when the ingoing path is true, it sets all the outgoing paths to be true |

However, UML activity diagram can only describe a high-level abstract model of the process. This is because of two problems. First is that UML does not consider the logical flow of execution among the activities. Second is that UML notations are very limited in number and meaning. Those problems reduce the overall understandability of a UML activity diagram. Accordingly, the UML activity diagram fails to provide an analytical model that can be used later by the developer as a guide during the implementation phase. The problems in UML have been been solved in the BPMN language [14]. BPMN provides the concept of control gateways. Therefore, it can precisely describe the logical follow of the activities within the process. Furthermore, BPMN provides rich set of notations, every notation has a predefined semantic, that accurately defines the meaning of this notation, this semantic is governed by a known syntax in case of combining two or more notations together. Table 1 summarizes and explains the meaning of the important BPMN notations that have been used to construct an analytical model of the proposed case study architecture.

#### B. Java Agent Development Framework

Java Agent DEvelopment (JADE) is a distributed middleware that is used to implement a Multi-Agent System (MAS) [16]. Fig. 3-a shows the deployment of the proposed MRS architecture via JADE Remote Monitoring Agent (RMA). Every component in the proposed architecture is represented as JADE software agent. Each JADE runtime instance is an independent thread that composes of a set of containers. A container is a group of agents run under the same runtime instance. Every runtime instance must have a main container with an Agent Management System (AMS) and a Directory Facilitator (DF). An AMS provides a unique Identifier (AID) that is used as a communication address for every agent. While, the DF announces the services that agents can offer. JADE complies with the Foundation for Intelligent Physical Agent (FIPA) specifications. FIPA is an IEEE Computer Society standards organization that promotes agent-based technology and the interoperability of its standards with other technologies. JADE agent uses FIPA-Agent Communication Language (FIPA-ACL) to exchange messages either inside or outside their runtime instance [17].

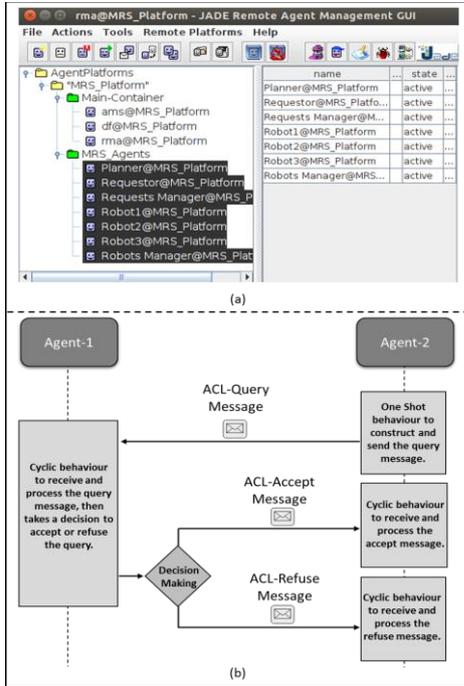

Fig. 3. (a) JADE framework – (b) JADE interaction diagram - an example

Every JADE agent applies a group of behaviors. A behavior is an event handler routine that is used by the agent to modify its parameters and negotiate with other agents as shown in Fig. 3-b. JADE offers different types of behaviors, four of those behaviors have been used during the implementation of this work, which are the following:

- One-shot behavior: a simple behavior that is executed once when it is called by the agent. Thus, it is useful to trigger an event and to send an ACL-Message.
- Cyclic behavior: a simple behavior that stays active as long as the agent is alive. Thus, it is so useful to receive a message with specific conversation-ID.
- Sequential behavior: a composite behavior that controls the sequence of execution of more than one-shot behavior. The sequence control is not always based on a fixed order, as it can be based on the sum of the input events from the agent environment.
- Parallel behavior: a composite behavior that concurrently controls the execution and the termination of more than one-shot behavior.

From the previous description of JADE framework, it can be seen that JADE is a proper tool to implement a BPMN model in the context of MRS. This is because of two reasons. First is that an analogy between BPMN and JADE concepts can be easily derived. An activity from the BPMN model can be coded as a simple one-shot or cyclic behavior in JADE. While, a gateway can be translated into a composite sequential or parallel behavior in JADE. Second, JADE implementation can be used as a simulator of the MRS, to test the different aspects of the system, the same way it is used in this article. Furthermore, the same implementation code can be used to deploy the system over real world hardware [18].

### C. Performance measurements

Measuring the MRS performance is a necessity to evaluate the system. The related research in [19] [20] proposes criteria such as distributability, diagnosability, modifiability, and modularity. However, these qualitative criteria are often relative and vague [21]. Thus, this article specifies absolute quantitative indicators and measures them during the runtime to evaluate the MRS performance. By thinking in the MRS as a black box that receives a number of requests, which either success or fail to be executed. The system performance can be expressed as a function in the following measurements:

- Throughput: the number of processed (successful and fail) requests per time unit.
- Latency: the time taken from the arrival of a request to the start of executing this request (shorter latency means better performance).
- Success rate: the number of successful requests divided by the number of received requests per time unit.
- Failure rate: the number of failed requests divided by the number of received requests per time unit.
- Efficiency: the success rate divided by the failure rate.

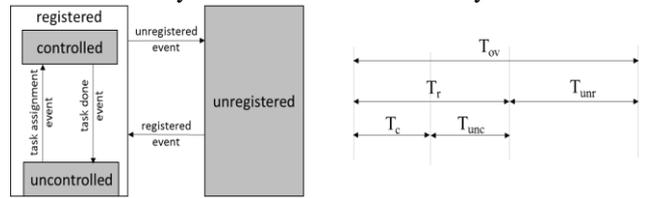

Fig. 4. robot state diagram

As the robots performance has a great influence on the MRS performance [22], it will be considered by this reach as well. The state diagram in Fig. 4 shows the different states that a robot can go through during its operation. Based on this diagram the following measurements can be obtained:

- Controlled time ($T_c$): the time taken by the robot to perform a task.
- Uncontrolled time ($T_{unc}$): the time that the robot is registered and waiting to be assigned for a task.
- Registered time ($T_r$): the summation of the robot controlled and uncontrolled time.
- Unregistered time ($T_{unr}$): the time that the robot is unregistered from the MRS.
- Overall time ($T_{ov}$): the summation of the robot registered and unregistered time.

Based on the above measurements, the following robot performance criteria can be calculated:

- Availability: the robot registered time ($T_r$) with respect to its overall time ($T_{ov}$).
- Utilization: the robot controlled time ($T_c$) with respect to its overall time ($T_{ov}$).
- Effectiveness: the robot controlled time ($T_c$) with respect to its uncontrolled time ($T_{unc}$).

## IV. SYSTEM MODEL

### A. Request Manager

The main responsibility of the requests manager is to receive the requests from different requestors, then find if there is an associated plan blueprint with this request in the KB, to send it to the planner for further processing. Fig. 5 shows the requests manager model as a BPMN activity diagram. In this diagram, the requests manager receives a request at the very beginning or during the process, and places it at the bottom of the request list. Then, it checks if there is another plan that has been sent for execution. If yes, the requests manager waits the plan execution feedback. If no, the requests manager selects the request at the top of the requests list to be processed. In this scenario, it has been assumed that the requests manager is scheduling the requests based on First Come First Serve (FCFS) method.

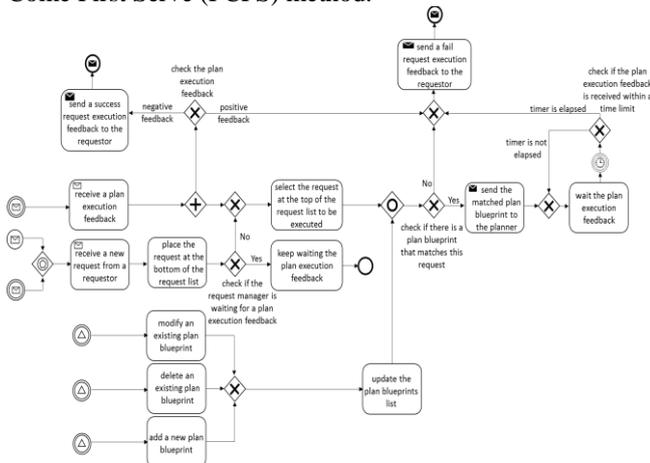

Fig. 5. requests manager activity diagram

The requests manager checks if the selected request matches any of the plan blueprints in the KB. Concurrently, the requests manager can add, delete, or modify the plan blueprints. Consequently, the requests manager handles any change in the plan blueprints during matching them with the selected request. If the requests manager does not find a plan blueprint match with the selected request, it sends a fail request execution to the requestor. If there is a match between one of the existing plan blueprints and the selected request, the requests manager sends the plan blueprint to the planner and initiates a time limit to receive a plan execution feedback. If the feedback is not received within this time limit, the requests manager considers this request is failed to be accomplished, and moves to the next request in the list. If the feedback is received within the time limit, the requests manager checks if this feedback is positive or negative, and sends a request success or fail response upon that.

*B. Planner*

The main responsibility of the planner is to receive a plan blueprint from the requests manager and checks if it can be transformed to a verified executable plan, by using the available robots. Fig. 6 shows the planner model as a BPMN activity diagram. In this diagram, when the planner receives a plan blueprint, it retrieves the current availability of the robots, their current capabilities, and their tasks history. Then it checks the number of registered robots. If it finds only one robot is available, it directly sends a fail plan execution to the requests manager, as it is known in advance that the plan needs more than one robot to be executed.

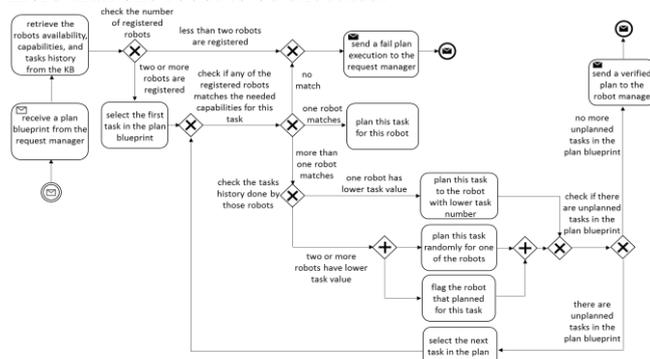

Fig. 6. planner activity diagram

If more than one robot is registered, the planner selects the first task in the plan blueprint, and compares it to the capabilities of the registered robot. If no robot matches with the capabilities that are needed to execute this task, the planner sends a fail plan execution to the requests manager. If one robot only matches, this robot will be planned to execute this task. Otherwise, if there are more than one robot that can perform this task, the planner tries to balance the tasks distribution among these robots. This is done by checking the tasks history of these robots, and then it plans this task for the robot with the lowest tasks history. Finally, if there are two or more robots with lowest tasks history, the planner plans the task for any of them randomly. Then it marks this robot, as if this case happens again, another robot will be planned for the task. After the planner plans a task for a specific robot, it goes through all the remaining tasks in the received plan blueprint, and performs the previously described algorithm. If all the tasks in the plan blueprint are planned. The planner sends this verified plan to the robots manager for execution.

*C. Robots Manager*

The main responsibility of the robots manager is to receive a verified plan from the planner and assign the tasks in this plan to the registered robots. Fig. 7 shows the robots manager model as a BPMN activity diagram. In this diagram, the robots manager receives the verified plan. Simultaneously, the robots manager is responsible for registering/unregistering the robots to/from the MRS. Consequently, the robots manager can handle any change in the robots availability during assigning the tasks in the plan. The robots manager assigns the first task in the plan to the associated robot for execution. Then, it initializes a time limit to receive a task execution feedback. If the feedback is not received within that time limit, the robots manager considers this plan as failed, and sends a plan execution fail feedback to the requests manager. If the feedback is received within the time limit, the robots manager checks if this feedback is positive or negative.

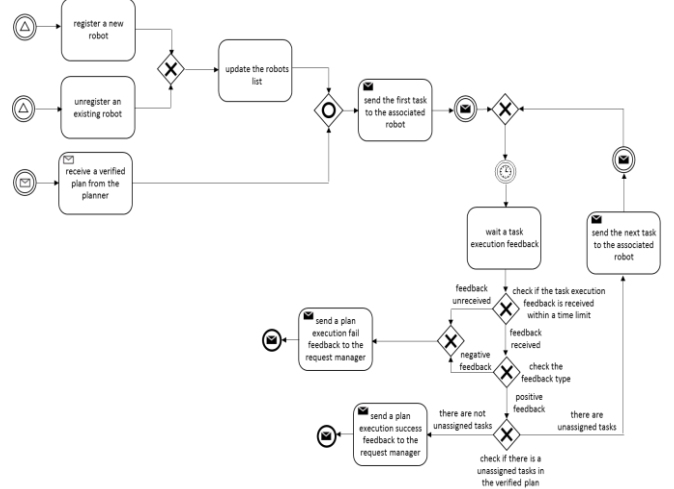

Fig. 7. robots manager activity diagram

If the feedback is negative, the robots manager considers this plan is failed, and sends a plan execution fail feedback to the requests manager. If the feedback is positive, the robots manager checks if there are remaining unassigned tasks in the verified plan. If so, it goes through them one by one, and performs the previously described algorithm. Finally, if all the tasks in the verified plan have been successfully executed, the robots manager sends a plan execution success feedback to the requests manager.

V. SYSTEM SIMULATION

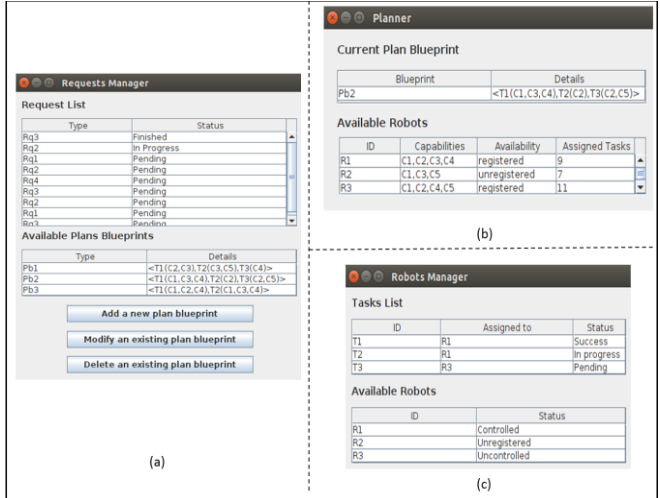

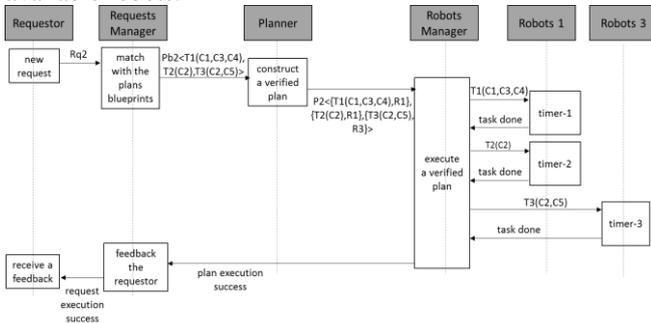

Fig. 8. simulation GUIs (a) requests manager – (b) planner – (c) robots manager

JADE is used to deploy the proposed MRS architecture. Each of the activity diagrams that discussed earlier is implemented as a JADE agent with a separate Graphical User Interface (GUI) as shown in Fig. 8. These GUIs are used to manipulate and monitor the agents (i.e., system components) parameters. Fig. 8-a shows the requests manager GUI that is used to add/remove/modify the plan blueprints, while the requests manager processes the requests. Fig. 8-b shows the planner GUI that is used to show the executed plan blueprint. Also, it shows the available robots and their status, capabilities, and tasks history. Fig. 8-c shows the robot manager GUI that is used to show the assigned tasks for the available robots.

Fig. 9. plan execution sequence diagram – an example

Fig. 9 is an example of the interaction among JADE agents to fulfil a request, while Fig. 10 shows the ACL-messages that are generated during this interaction. The example explains the execution of request ($Rq_2$) that can be seen in the requests manager GUI in Fig. 10-a. The interaction starts when $Rq_2$ status is in progress. $Rq_2$ has matched a plan blueprint ($Pb_2$). The requests manager sends the contents of the plan blueprint to the planner in an ACL-massage as shown in Fig. 10-a. The planner constructs a verified plan ($P_2$) by distributing the tasks over the available robots based on their capabilities and tasks history. At this time, only two robots are registered which are $R_1$ and $R_3$. $R_1$ has done 9 tasks in its task history, while $R_3$ has done 11 tasks in its task history, as shown in Fig. 10-b.

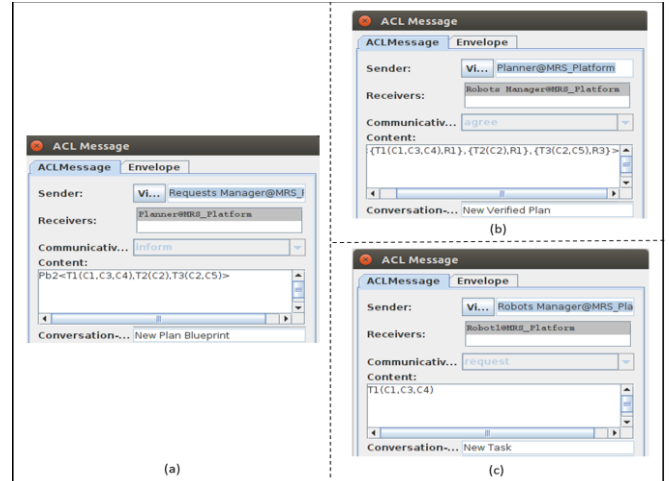

Fig. 10. ACL-messages contain (a) plan blueprint – (b) verified plan – (c) task

Task ($T_1$) requires capabilities ($C_1$, $C_3$, $C_4$), which are unique capabilities of $R_1$, therefore $T_1$ is planned to $R_1$. $T_2$ requires ($C_2$), which is a common capability between $R_1$ and $R_3$. Therefore, the planner checks that $R_1$ has 9 tasks in its history plus a planned task $T_1$. This is less than the task history of $R_3$ which is 11 tasks. Therefore, the planner decides to plan this task to $R_1$, to balance the tasks among the robots. $T_3$ requires ($C_2$, $C_5$) which are unique tasks of $R_3$, therefore $T_3$ is planned to $R_3$. The planner sends the verified plan to the robots manager in an ACL-message as shown in Fig. 10-b. The robots manager parses the contents of the received ACL-messages, then it starts to assign the tasks one by one based on the verified plan. The task assignment is also done by sending an ACL-message as shown in in Fig. 10-c. The robot agent simulates a task by triggering a random timer. When the timer is elapsed, the robot agent sends a task done feedback to the robots manager. When the robots manager is sure that all the tasks are done, it sends a plan execution success feedback to the requests manager.

## VI. SYSTEM PERFORMANCE ANALYSIS

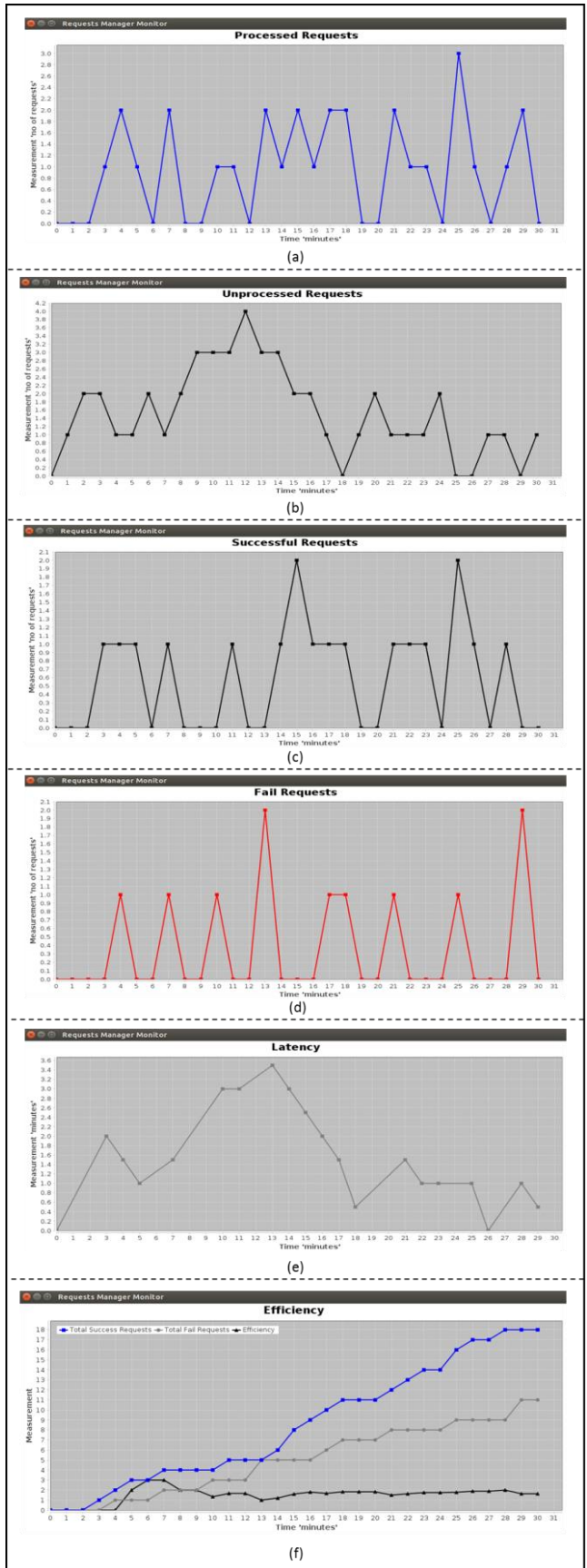

Fig. 11. (a) processed requests – (b) unprocessed requests – (c) successful requests – (d) fail requests – (e) latency – (f) efficiency

As mentioned earlier, the three sources of variations in the proposed case study are the plan blueprints, the number of registered robots, and the capabilities of every robot. The number of registered robots has been selected to vary during the system runtime. Therefore, the system performance can be measured. The system simulation over JADE has been run for 30 minutes. During every minute of the runtime, the requestor agent generates a random request. Additionally, one of the three robots agents randomly unregistered, and one randomly registered. The plan blueprints and the robots capabilities are fixed through the runtime.

The first set of performance measurements that are shown in Fig. 11 are calculated by the requests manager, because they essentially depend on monitoring the requests status. The processed requests chart in Fig. 11-a shows how many requests have been processed by the system at a certain time. Therefore, this chart expresses the system throughput (i.e., how fast is the system). The system throughput cannot be understood as an absolute value, because it is the summation of successful and fail requests. For example at the 4$^{th}$ minute of Fig. 11-a, the number of processed requests are two. One of these requests was successful and the other was fail, as it can be seen in Fig. 11-c and Fig. 11-d respectively. The system latency is shown in Fig. 11-e. The system latency is a very important performance measurement, as it expresses how much the system is delayed or late. For example at the 26$^{th}$ minutes of Fig. 11-e, the latency is zero. This means that the system has no time delay in processing the incoming requests at this moment. The system latency is a function of the unprocessed requests history, that can be seen in Fig. 11-b. The system efficiency is obtained by dividing the accumulated number of successful request by the accumulated number of the failed request. Therefore, when the system efficiency is below one, this means that the system is inefficient. However, Fig. 11-f shows that the efficiency value was higher than one most of the runtime.

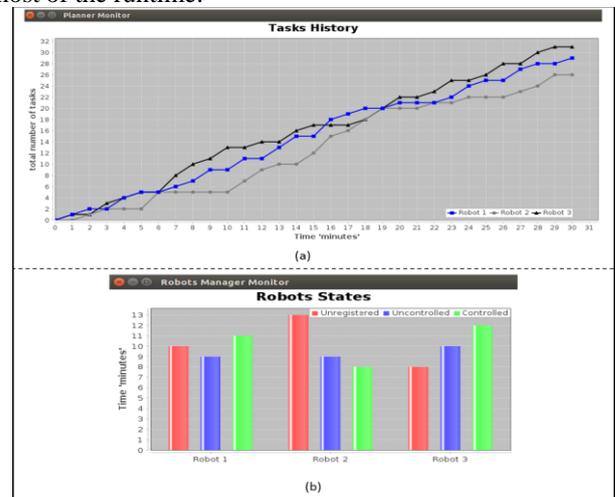

Fig. 12. (a) robots tasks history – (b) robots states

TABLE II. ROBOTS AVAILABILITY, UTILIZATION, AND EFFECTIVENESS

| | Robot 1 (R1) | Robot 2 (R2) | Robot 3 (R3) |
|---|---|---|---|
| Availability $\frac{T_r}{T_{ov}}$ | $\frac{20 \text{ min}}{30 \text{ min}} = 0.67$ | $\frac{17 \text{ min}}{30 \text{ min}} = 0.57$ | $\frac{22 \text{ min}}{30 \text{ min}} = 0.73$ |
| Utilization $\frac{T_c}{T_{ov}}$ | $\frac{11 \text{ min}}{30 \text{ min}} = 0.37$ | $\frac{8 \text{ min}}{30 \text{ min}} = 0.27$ | $\frac{12 \text{ min}}{30 \text{ min}} = 0.4$ |
| Effectiveness $\frac{T_c}{T_{unc}}$ | $\frac{11 \text{ min}}{10 \text{ min}} = 1.1$ | $\frac{8 \text{ min}}{9 \text{ min}} = 0.89$ | $\frac{12 \text{ min}}{10 \text{ min}} = 1.2$ |

Fig. 12-a shows the robots tasks history that is calculated by the planner. Although that the robots availability is changing during the runtime as shown in Fig. 12-b, the planner tries to balance the distribution of the number of tasks among the robots. For example, at the 6$^{th}$ minute in Fig. 12-a, the planner successes to balance all the robots tasks at the value of 5 tasks per each. Then the robots tasks history started to diverge, however the planner successes again to balance the robots tasks again at the 20$^{th}$ minute at the value of 20 tasks per each. Fig. 12-b shows the distribution of the robots states over the runtime period. Fig. 12-b is calculated by the robots manager as it directly monitors the robots status. Table 2 can be calculated based on the values of Fig. 12-b. It can be seen from Table 2 that $R_3$ is the most effective robot, as it was the most available and therefore the most utilized. $R_1$ and $R_2$ are less effective than $R_3$, however it can be seen that the planner tries to maximize their utilization, to reach to a close value of $R_3$ utilization.

## VII. SUMMARY, DISCUSSION, AND FUTURE WORK

This article has highlighted a new dimension of the MRS problem, which is modeling and simulation of the solution architecture during the design phase, and hence evaluating its performance. Therefore, this approach can be used to compare different system architectures to find the best solution based on the system requirements. Modeling an MRS architecture enables the common understanding and the critical thinking of the system among the development team. Furthermore, it enables the separation between the conceptual model and the implemented technology. An ADL such as UML fails to clearly represent the process model of the MRS components, because UML lacks of the notations that express the logical relation between the tasks in a process model. Therefore, the BPMN language is used to provide highly detailed analytical models of the components of the proposed architecture. Those analytical models are used during the coding to implement the different components of the system architecture. JADE agent development middleware is used to implement the MRS components, as the dynamic interaction among those components is needed to simulate the overall behavior. However, other technologies such as Robot Operation System (ROS) or Web Service (WS) can also be used to implement the very same models.

The simulation model has been used to measure a group of performance indicators under predefined constrains. Thus, these indicators are used to analysis and evaluate the proposed system architecture. Other qualitative criteria such as fault-tolerance or robustness can also be tested during the design phase by using the same simulation model. This can be done by replicating or switching off some of the model components, then studying their effect on the system performance. Although the MRS scheduling and planning is not the main goal of this research, our solution approach can be used to simulate different planning or scheduling methods, to find the best technique during the design phase.

As the BPMN is originally designed to describe the business processes, an extension of this notation language is required to represent the physical layer of the MRS. This extension should reflect a better image of the humans and robots activities as the main elements of the MRS. Additionally, the BPMN provides a mapping to the Business Process Execution Language (BPEL), which is an executable language that orchestrates the information among the WSs. Therefore in the future work, the same idea of constructing an executable model that can directly be used to generate the code will be considered. This can dramatically reduce the coding time and effort. Also in the future work, the same performance measurements that have been used in this article can be used in the implementation phase, as a part of the system visualization.